\documentclass{article}
\usepackage{arxiv}  

\usepackage{cite}
\usepackage{amsmath,amssymb,amsfonts}
\usepackage{algorithmic}
\usepackage{graphicx}
\usepackage{textcomp}
\usepackage{bmpsize}
\usepackage{xcolor}
\usepackage{lipsum}
\usepackage{hyperref}

\def \SAD {DA\textsuperscript{3}D}
\def \AAA {A\textsuperscript{3}}

\usepackage{caption}
\usepackage{subcaption}
\usepackage{tikz}
\usepackage{pgfplots}
\pgfplotsset{compat=1.16}

\usepackage{amsmath,amssymb,amsfonts}
\usepackage{cleveref}
\usepackage{enumitem}

\usepackage{makecell}
\usepackage{booktabs}
\usepackage{rotating}
\usepackage{tablefootnote}

\usepackage{algorithm2e}
\SetKwComment{Comment}{// }{}

\usepackage{pgfkeys,pgfmath}  
\usepackage{siunitx}
\usepackage{expl3}
\ExplSyntaxOn
\newcommand{\fpabs}[1]{\fp_eval:n{abs(#1)}}
\ExplSyntaxOff
\newcommand{\percincrease}[3][0]{%
    \pgfmathdivide{#3}{#2}%
    \pgfmathparse{\pgfmathresult-1}%
    \pgfmathmultiply{\pgfmathresult}{100}%
    \SI[round-mode=places,round-precision=#1]{\pgfmathresult}{\percent}
}%

\usepackage{abbr}
\usepackage{style}

\newcommand\copyrighttext{%
  \footnotesize \textcopyright 2022 IEEE. Personal use of this material is permitted. Permission from IEEE must be obtained for all other uses, in any current or future media, including reprinting/republishing this material for advertising or promotional purposes, creating new collective works, for resale or redistribution to servers or lists, or reuse of any copyrighted component of this work in other works.}
\newcommand\copyrightnotice{%
\begin{tikzpicture}[remember picture,overlay]
\node[anchor=south,yshift=2cm] at (current page.south) {\fbox{\parbox{\dimexpr\textwidth-\fboxsep-\fboxrule\relax}{\copyrighttext}}};
\end{tikzpicture}%
}

\begin{document}

\title{Double-Adversarial Activation Anomaly Detection:\\Adversarial Autoencoders are Anomaly Generators}

\author{
	Jan-Philipp Schulze \\
	Fraunhofer AISEC \\
	\texttt{jan-philipp.schulze\thanks{\texttt{@aisec.fraunhofer.de}}} \\
	\And
	Philip Sperl \\
	Fraunhofer AISEC \\
	\texttt{philip.sperl\footnotemark[\value{footnote}]} \\
	\And
	Konstantin B\"ottinger \\
	Fraunhofer AISEC \\
	\texttt{konstantin.boettinger\footnotemark[\value{footnote}]} \\
}

\maketitle
\copyrightnotice

\begin{abstract}
Anomaly detection is a challenging task for machine learning methods due to the inherent class imbalance. 
It is costly and time-demanding to manually analyse the observed data, thus usually only few known anomalies if any are available.
Inspired by generative models and the analysis of the hidden activations of neural networks, we introduce a novel unsupervised anomaly detection method called \SAD{}.
Here, we use adversarial autoencoders to generate anomalous counterexamples based on the normal data only.
These artificial anomalies used during training allow the detection of real, yet unseen anomalies.
With our novel generative approach, we transform the unsupervised task of anomaly detection to a supervised one, which is more tractable by machine learning and especially deep learning methods.
\SAD{} surpasses the performance of state-of-the-art anomaly detection methods in a purely data-driven way, where no domain knowledge is required.


\end{abstract}

\section{Introduction}
In anomaly detection (AD), we analyse data for samples that deviate from the given notion of normal.
Based on the use case, these anomalies could lead to e.g. attacks on infrastructure, fraudulent behaviour or manufacturing errors.
AD is usually seen as an unsupervised task, where we do not have any prior knowledge about the training data, but consider the majority of samples to be normal.
On the other hand, semi-supervised AD methods (e.g. \cite{ruff_deep_2020,pang_deep_2019,sperl_activation_2021}) incorporate known anomalies.
Whereas unsupervised AD methods are widely applicable, semi-supervised methods excel in performance the more anomalies are available.
In \SAD{}, we combine both research directions to allow the application of reliable semi-supervised AD approaches in unsupervised settings:
by analysing the normal data only, we generate artificial anomaly samples.
This task is challenging as the generated anomalies must be realistic enough to aid the training process, but may not be too close to normal ones to decrease the performance.

Despite their superior performance in common machine learning (ML) tasks, it is still challenging to apply deep learning (DL) methods to AD.
AD is characterised by its inherent class imbalance, where normal samples are abundant but finding anomalies may require significant manual work and domain expert knowledge.
DL methods learn to select the most relevant parts of the input given their objective function and thus are especially data-demanding.
Generative adversarial nets \cite{goodfellow_generative_2014} (GANs) are a DL-based method for data distribution estimation.
Closely related, adversarial autoencoders \cite{makhzani_adversarial_2016} (AAEs) reconstruct the given input under a certain prior distribution in the latent space.
We adapt AAEs in a novel way to generate artificial anomalous data points.
Generating artificial anomalies may seem counter-intuitive at first.
Note that we do not need to create anomalies that resemble real anomalies.
In fact, we try to generate samples that allow DL algorithms to find a suitable decision boundary that generalises to real data.
Figuratively, our method recombines patterns seen in the normal training data.
Even if real anomalies are not a combination of normal patterns, we show that our artificial anomalies are useful to separate normal data from other, i.e. anomalous, data points.
Throughout our research, we profited from the insights gained by a semi-supervised AD method called \AAA{} \cite{sperl_activation_2021}.
Thanks to their end-to-end neural architecture, we use \AAA{} as a building block in \SAD{}.
We do so by introducing two adversarial objectives leading to non-trivial artificial anomalies.
Based on this principle, we call our new unsupervised AD method \SAD{}: \underline{d}ouble-\underline{a}dversarial \underline{a}ctivation \underline{a}nomaly \underline{d}etection.

In summary, we make the following contributions:
\begin{itemize}[topsep=0pt,noitemsep]
    \item We introduce \SAD{}, a data-driven GAN-based unsupervised AD method analysing the activations of AAEs.
    \item We describe a method to generate non-trivial synthetic anomalies, which are useful to detect real anomalies.
    \item We evaluate \SAD{} on ten data sets against six baseline methods and plan to open-source our code.
\end{itemize}

\subsection{Related Work}
AD profits from a wide range of research \cite{ruff_unifying_2021}, featuring a variety of underlying ML algorithms.
Widely acknowledged unsupervised approaches use one-class classifiers, e.g. OC-SVMs \cite{scholkopf_support_2000}, or distance metrics, e.g. Isolation Forest \cite{liu_isolation_2008}.

\paragraph{DL-based Anomaly Detection}
Over the past years, the popularity of DL-based AD methods \cite{pang_deep_2021} has grown.
Popular ideas use autoencoders (AEs), which are a special type of NN reconstructing its input under small hidden dimensions.
Here, the reconstruction error \cite{borghesi_anomaly_2019} may be used for AD or the AE is a building block in a more complex pipeline as in DAGMM \cite{zong_deep_2018}.
Whenever anomalies have been found manually, semi-supervised methods try to increase the overall detection performance by incorporating this prior knowledge, e.g. in DevNet \cite{pang_deep_2019}, DeepSAD \cite{ruff_deep_2020} or \AAA{} \cite{sperl_activation_2021}.
\SAD{} works in the strictest AD setting: a fully unsupervised one as often found in practice, where it is infeasible to manually analyse the data for anomalies.

\paragraph{GAN-based Anomaly Detection}
Since the introduction of GANs, multiple concepts have been developed to apply their potential in AD.
A commonly used idea is measuring the distance between the input and the generated samples given a generator trained on normal samples only.
The reconstruction error is used in e.g. AnoGAN \cite{schlegl_unsupervised_2017,schlegl_f-anogan_2019}, ALAD \cite{zenati_adversarially_2018}, MAD-GAN \cite{li_mad-gan_2019} or GANomaly \cite{akcay_ganomaly_2019}.
Recent studies have shown that combining several GANs at once may further boost the AD performance \cite{han_gan_2021}.
Measuring the reconstruction errors of AAEs has also been proposed \cite{vu_anomaly_2019}, refined by optimisations in presence of outliers \cite{beggel_robust_2020}.
Our method is not based on reconstruction distances, which have known drawbacks:
subtle changes are hard to detect and anomalies, which are close to the normal samples, may still be reconstructed well.
Instead, we use the principles of activation analysis \cite{sperl_activation_2021,raghuram_general_2021}, where all hidden activations of an NN are analysed at once.
GANs comprise a generator and a discriminator network, where the former tries to generate samples that align with the training samples.
In AD research, the generator was used to either generate new normal samples based on known normal ones, e.g. in DOPING \cite{lim_doping_2018}, or to generate new anomalous samples using known anomalies \cite{salem_anomaly_2018}.
\SAD{} is an unsupervised method, thus does not have access to anomalous samples.
To further improve AD, we generate anomalous samples from normal samples only.
There is prior work on generating corner cases of the normal data \cite{bhatia_exgan_2021}, or generating anomalies in behavioural patterns \cite{zheng_one-class_2019}.

\paragraph{Anomaly Detection based on Synthetic Anomalies}
We generate anomalous samples from normal samples only, sometimes referred to ``negative sampling'' or ``synthetic anomalies''.
For some data types, e.g. images, simple counterexamples may be found by geometric transformations \cite{golan_deep_2018} or out-of-distribution data \cite{hendrycks_deep_2019}.
On tabular data, linear transformations \cite{bergman_classification-based_2020} are useful to generate counterexamples.
These manual transformations need careful adaptation to the respective data.
In contrast, we designed \SAD{} as a data-driven method applicable to several use cases at once.
We do so by sampling synthetic anomalies from the latent space of AAEs.
Sipple \cite{sipple_interpretable_2020} and Somepalli~et~al. \cite{somepalli_unsupervised_2021} have shown that some regions may be useful to generate synthetic anomalies.
In \SAD{}, we improve on this manual analysis with a generative model, which automatically decides which regions are useful to improve the AD performance.
Our adversarial goals refine the synthetic anomalies while the anomaly detector learns to distinguish them from normal samples, thus improving the generator and detector at once.
FenceGAN \cite{ngo_fence_2019} is the closest to our method.
Here, the authors generate samples at the boundary of normal instances.
Finding samples at the boundary implies a significant data overhead as both, normal and anomalous samples are needed to evaluate suitable parameters.
Moreover, the data might contain multiple separate notions of normal, where it is hard to find a common border \cite{diakonikolas_complexity_2020}.
We introduce \SAD{} as a data-driven AD method working in a variety of applications without further adaptations.

\section{Prerequisites}
We define NNs as a function $f_\text{NN}(\vect{x}; \vectg{\theta}) = \hat{\vect{y}}$ approximating the mapping from the input $\vect{x}$ to the output $\hat{\vect{y}}$ under the parameters $\vectg{\theta}$.
As abbreviation, we write $f_\text{NN}: \vect{x} \mapsto \hat{\vect{y}}$.
The hidden layers give rise to the activation $\vect{h}_i$.
When analysing all hidden activations, we write $[\vect{h}_i]_i = [\vect{h}_0, \vect{h}_1, \ldots]$.
Let $a(f)=[\vect{h}_{f, i}]_i$ be the function that extracts the activations from an NN $f$.

\subsection{GANs \& AAEs}
\SAD{} uses the principles introduced in Generative Adversarial Nets \cite{goodfellow_generative_2014} and Wasserstein GANs \cite{arjovsky_wasserstein_2017}.
GANs comprise two NNs: the generator and the discriminator.
Given a sample from the training set $\vect{x} \in \set{X}$, the generator, $f_\text{gen}: \vect{n} \mapsto \tilde{\vect{x}}$ generates samples that look indistinguishable from these, i.e. follow the same distribution $\tilde{\vect{x}} \sim P_\set{X}(\vect{x})$.
The generator takes Gaussian noise as input, $\vect{n} \sim \set{N}(\vect{0}, \vect{1})$.
During training, the discriminator distinguishes between real and generated inputs.
The generator's objective is to fool the discriminator, which in turn adapts to the features of the improved generated images.
Adversarial autoencoders \cite{makhzani_adversarial_2016} apply the principle of GANs to AEs.
Like usual AEs, they comprise two parts: the encoder and the decoder network.
The encoder summarises the input to a low dimensional code vector, $f_\text{enc}: \vect{x} \mapsto \vect{h}_\text{code}, \dim{(\vect{x})} > \dim{(\vect{h}_\text{code})}$.
In turn, the decoder reconstructs the input from the code vector, $f_\text{dec}: \vect{h}_\text{code} \mapsto \hat{\vect{x}}$.
During training, a discriminator distinguishes between a prior distribution and the code vector.
As consequence, the training samples will e.g. follow a Gaussian distribution in the code layer.

\subsection{Activation Analysis}
\SAD{} profits from the semi-supervised AD method \AAA{} \cite{sperl_activation_2021}.
\AAA{} combines two NNs: the target and the alarm network.
The authors show that the activations of the target network are different for samples that it was trained on, i.e. normal ones, and others, i.e. anomalous ones.
The alarm network analyses these activations.
During training, the authors use all normal samples, a few known anomalies and trivial anomalies, i.e. Gaussian noise.
We believe that the simple type of artificial anomalies limits \AAA{} to semi-supervised settings.
For \SAD{}, we transfer the concept of activation analysis to unsupervised AD by introducing an anomaly generator based on AAEs.

\begin{figure}[tb]
  \begin{center}
    \input{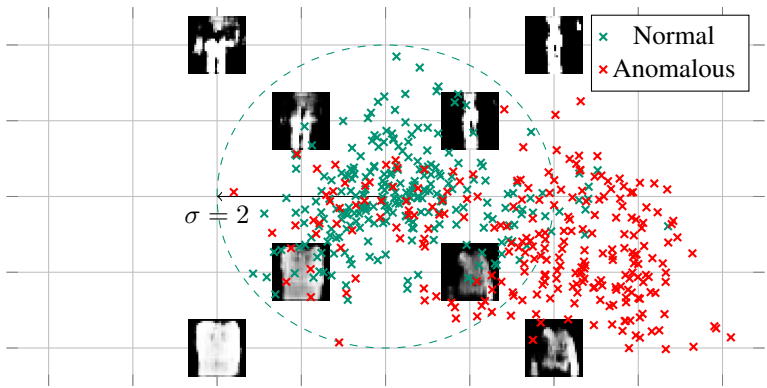}
    \caption{
        The first 2 latent space dimensions of a 16-dimensional AAE trained on FMNIST.
        We defined the first 5 classes to be normal, the rest anomalous.
        Normal samples lie close to the centre, whereas anomalies are likely outside.
        Following this intuition, we generate synthetic anomalies.
    }
    \label{fig:geometric}
  \end{center}
\end{figure}

\section{\SAD{}}
\SAD{} explores the following intuition as motivated in \cref{fig:geometric}:
\begin{quote}
    \label{assumption}
    Let $f_\text{AAE} = f_\text{dec} \circ f_\text{enc}$ be an AAE trained on normal data only.
    Due to the normalisation during training, most normal samples lie within $\set{N}(\vect{0},\vect{1})$ in the code layer.
    Thus, we can generate anomalous counterexamples by sampling from regions outside of the normal clusters.
    These artificial anomalies can be used to detect real anomalies.
\end{quote}

As analogy, imagine the AAE's decoder to be a machine generating realistic images.
Based on the training data the operator has learned which features, e.g. certain painting strokes, form the images.
In \SAD{}, we decide which buttons to press to generate anomalies.
Although these anomalies are combinations of normal patterns, we show that they are useful to find a decision boundary between real normal and real anomalous samples.

\subsection{Architecture}
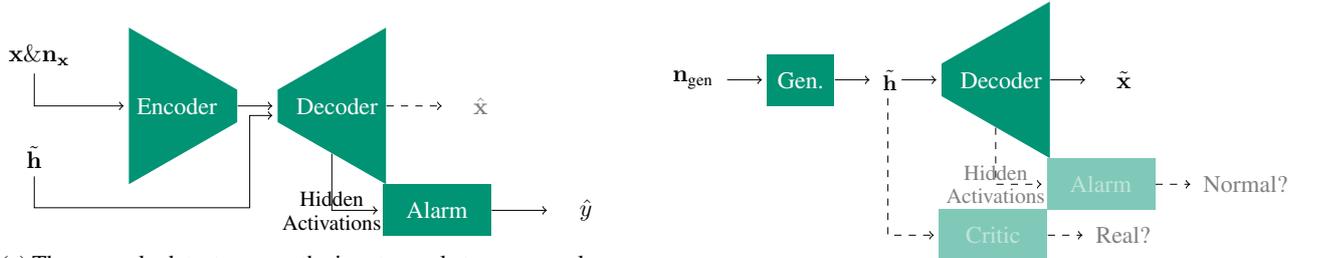
\begin{figure}[tb]
     \centering
     \begin{subfigure}[b]{0.475\textwidth}
         \centering
         \scalebox{.9}{\begin{tikzpicture}[node distance=1.45 * \NodeSepX cm]

	\node[encoder] (ae) {Encoder};
	\node[decoder, right of=ae] (dec) {Decoder};

	\draw[arrow] (ae) -- (dec);

	\node[smallannot, left of=ae, yshift=\NodeSepY cm] (x) {$\vect{x} \& \vect{n}_\vect{x}$};
	\node[smallannot, left of=ae, yshift=-\NodeSepY cm] (xt) {$\tilde{\vect{h}}$};

	\draw[arrow] (x) |- (ae);
	\draw[arrow] (xt) |- +(2.1*\NodeSepX cm, -\NodeSepY cm) |- (dec.190);

	\node[smallannot, right of=dec, gray] (x_pred)  {$\hat{\vect{x}}$};

	\draw[arrow, dashed] (dec) -- (x_pred);

	\node[rect, below right of=dec] (alarm) {Alarm};

	\draw[arrow] (dec) |-  node[tinyannot, text width=\NodeSepX cm] {Hidden \\ Activations} (alarm);

	\node[smallannot, right of=alarm] (out) {$\hat{y}$};
	\draw[arrow] (alarm) -- (out);

	
\end{tikzpicture}}
         \caption{
            The anomaly detector maps the input sample to an anomaly score $\hat{y}$.
            During training, the alarm network analyses the hidden activations in the AAE caused by the normal data and synthetic anomalies.
        }
         \label{fig:detector}
     \end{subfigure}
     \hfill
     \begin{subfigure}[b]{0.475\textwidth}
         \centering
         \scalebox{.9}{\begin{tikzpicture}[node distance=1.45 * \NodeSepX cm]

	\node[rect, minimum width=.5*\NodeSepX cm, text width=.5*\NodeSepX cm] (enc) {Gen.};
	\node[tinyannot, right of=enc, xshift=-.6*\NodeSepX cm] (code) {$\tilde{\vect{h}}$};
	\node[decoder, right of=code, xshift=-.4*\NodeSepX cm] (dec) {Decoder};
	\node[smallannot, left of=enc, xshift=.4*\NodeSepX cm] (x) {$\vect{n}_\text{gen}$};
	\node[smallannot, right of=dec, xshift=-.2*\NodeSepX cm] (y) {$\tilde{\vect{x}}$};

	\node[rect, below right of=code, opacity=.5, yshift=-\NodeSepY cm] (disc) {Critic};
	\node[smallannot, right of=disc, opacity=.5, xshift=-.2*\NodeSepX cm] (yd) {Real?};

	\node[rect, below right of=dec, opacity=.5] (AD) {Alarm};
	\node[smallannot, right of=AD, opacity=.5, xshift=-.2*\NodeSepX cm] (normal) {Normal?};

	\draw[arrow] (x) -- (enc);
	\draw[arrow] (enc) -- (code);
	\draw[arrow] (code) -- (dec);
	\draw[arrow] (dec) -- (y);

	\draw[arrow, dashed] (code) |- (disc);
	\draw[arrow, dashed] (disc) -- (yd);

	\draw[arrow, dashed] (dec) |- node[tinyannot, text width=\NodeSepX cm, opacity=.5] {Hidden \\ Activations} (AD);
	\draw[arrow, dashed] (AD) -- (normal);


\end{tikzpicture}}
         \caption{
            The anomaly generator maps noise to non-trivial anomalies $\tilde{\vect{h}}$.
            Based on the feedback of the critic and alarm network, it refines the synthetic anomalies.
        }
         \label{fig:generator}
     \end{subfigure}
        \caption{Architecture of the anomaly detector, $f_\text{AD}$, and anomaly generator, $f_\text{AG}$.}
        \label{fig:arch}
\end{figure}

\SAD{} follows a feedback-driven architecture featuring two parts: the anomaly detector $f_\text{AD}$ and the anomaly generator $f_\text{AG}$.
The anomaly detector maps the input sample to an anomaly score, $f_\text{AD}: \vect{x} \mapsto \hat{\vect{y}} \in [0,1]$, where 1 is highly anomalous.
During training, it receives artificial anomalies produced by the generator.
When evaluating new samples, only $f_\text{AD}$ is used.
Although similar to the discriminator-generator pair of GANs, the task differs:
we generate anomalous samples, but only have a notion of normal ones.
This is made possible by a discriminator that is involved in the AD task: the AD method itself.
Throughout this chapter, we use three types of input data:
\begin{enumerate}[topsep=0pt,noitemsep]
    \item 
    A \emph{training sample}, $\vect{x}$, from the training set $\set{X}$,
    \item
    A \emph{trivial anomaly}, $\vect{n}_\vect{x}$, in the input space from a Gaussian distribution,
    \item
    A \emph{generated anomaly}, $\tilde{\vect{h}}$, in the latent space using \SAD{}'s anomaly generator.
\end{enumerate}
Trivial anomalies were introduced in \AAA{}:
as the real inputs are scaled to $[0,1]$, trivial anomalies use a Gaussian prior $\vect{n}_\vect{x} \sim \set{N}( \vect{0.5}, \vect{1})$.
In our work, we expand the artificial anomalies to the latent space of AAEs.
Here, the normal samples are normalised, thus we generate anomalies by sampling around them.
We show our architecture in \Cref{fig:arch} and discuss the two components in the following.

\subsection{Anomaly Detector, $f_\text{AD}$}
\label{ad:detect}

The anomaly detector maps a given input to an anomaly score.
It comprises three components as shown in \Cref{fig:detector}:
1) the AAE's encoder,
2) the AAE's decoder
and 3) the alarm network.
The alarm network analyses the activations of the decoder and outputs an anomaly score:
$
f_\text{AD} = f_\text{alarm} \circ a(f_\text{dec}) \circ f_\text{enc}: \vect{x} \mapsto \hat{\vect{y}}
$.
Whereas the normal training data should be detected as normal, trivial and generated anomalies should be anomalous.
During training, only the mapping parameters of the alarm network $\theta_\text{alarm}$ are adapted, while the encoder and decoder remain in their pretrained state:
\begin{equation}
\label{eq:alarm}
\begin{split}
\argmax_{\theta_\text{alarm}}
\Expected
_{
\vect{x}, \vect{n}_\vect{x}, \tilde{\vect{h}}
}
[
    \underbrace{\log(1 - f_\text{AD} ( \vect{x} ) )}_{\text{normal data}}
    + \underbrace{\log(f_\text{AD} ( \vect{n}_\vect{x} ) )}_{\text{trivial anom.}}
    + \underbrace{\log(f_\text{alarm} ( a( f_\text{dec} ( \tilde{\vect{h}} )))) )}_{\text{generated anom.}}
].
\end{split}
\end{equation}
Remember that $a(\cdot)$ extracts the activations of the underlying NN.
We expect these activation patterns to differ for normal samples and our synthetic anomalies.

\subsection{Anomaly Generator, $f_\text{AG}$}
\label{ad:gen}

The anomaly generator outputs artificial anomalies in the AAE's code space, which are used during the training of the anomaly detector.
We evaluate two types of anomaly generator:
a) a simple Gaussian prior, and b) an advanced GAN-based one as explained in the following.
Our anomaly generator comprises four components as shown in \Cref{fig:generator}:
1) the generator,
2) the critic,
3) the alarm network
and 4) the AAE's decoder.
Overall, the anomaly generator becomes:
$
f_\text{AG} = f_\text{dec} \circ f_\text{gen}: \vect{n} \mapsto \tilde{\vect{h}} \mapsto \tilde{\vect{x}}
$.

Trivial anomalies alone do not carry enough information to improve the detection performance; however, too realistic samples may be confused with normal inputs, thus increasing the false positive rate.
We balance between these goals by imposing two adversarial objectives:
\begin{enumerate}[noitemsep]
    \item \emph{Fool the anomaly detector}:
    Generate samples that are falsely classified as normal.
    Encourages more subtle anomalies as the detector improves.
    \item \emph{Ignore known normal samples}:
    Discourage imitating the training samples.
    Forces the anomaly generator to explore new regions to form the outputs.
\end{enumerate}
Inspired by the learning process of GANs, we enforce both objectives by competing NNs.
Whereas goal~1 opposes the anomaly detector, we introduce a separate critic NN for goal~2:
Remember that the AAE bounds the training samples within a unit-variance Gaussian distribution in the code layer.
We thus use a wider Gaussian distribution for the critic, $\vect{n}_\vect{h} \sim \set{N}(\vect{0}, \vect{4})$.
\begin{equation}
\label{eq:critic}
\begin{split}
\argmin_{\theta_\text{critic}}
\Expected_{
\vect{x}, \vect{n}_\text{gen}, \vect{n}_\vect{h}
}
[
\underbrace{f_\text{critic} \left( \vect{n}_\vect{h} \right)}_{\text{stay in boundaries}}
- \lambda_\text{g} \underbrace{f_\text{critic} \left( f_\text{gen} \left( \vect{n}_\text{gen} \right) \right)}_{\text{incentivise exploration}}
- \lambda_\text{n} \underbrace{f_\text{critic} \left( f_\text{enc} \left( \vect{x} \right) \right)}_{\text{avoid normal}}
],
\end{split}
\end{equation}
where the weighting factors were set to $\lambda_\text{n} = \lambda_\text{g} = 2$ to penalise the generation of normal samples more.
The generator maps a prior to non-trivial anomalies.
During training, the anomaly generator fools the anomaly detector and the critic.
Similar to other GANs, we use a unit-variance Gaussian distribution, $\vect{n}_\text{gen} \sim \set{N}(\vect{0}, \vect{1})$:
\begin{equation}
\label{eq:gen}
\begin{split}
\argmin_{\theta_\text{gen}}
\Expected_{
\vect{n}_\text{gen}
}
[
 \underbrace{
 f_\text{alarm} \left(
 a \left(
 f_\text{dec} \left(
 f_\text{gen} \left(
 \vect{n}_\text{gen}
 \right)
 \right)
 \right)
 \right)
 }_{\text{fool the anomaly detector}}
 +  \underbrace{f_\text{critic} \left( f_\text{gen} \left( \vect{n}_\text{gen} \right) \right)}_{\text{fool the critic}}
].
\end{split}
\end{equation}

\begin{algorithm}[tb]
\caption{High-level overview about the data flow while training \SAD{}.}\label{alg:training}
\KwIn{$f_\text{AAE}, \set{X}_\text{train}$}
\KwResult{$f_\text{AD}$}
\For{$\vect{x}_\text{batch} \in \set{X}_\text{train}$}{
    \Comment{Generate new training samples}
    $\vect{n}_\vect{x} \gets \set{N}(\vect{.5}, \vect{1}),$
    $\vect{n}_\vect{h} \gets \set{N}(\vect{0}, \vect{4}),$
    $\vect{n}_\text{gen} \gets \set{N}(\vect{0}, \vect{1})$\;
    
    $\tilde{\vect{h}} \gets f_\text{AG}(\vect{n}_\text{gen})$\;
    
    \Comment{Train anom. detector \& generator}
    $f_\text{alarm} \gets (\vect{x}_\text{batch}, 0), (\vect{n}_\vect{x}, 1), (\tilde{\vect{h}}, 1)$, \cref{eq:alarm}\;    
    $f_\text{gen} \gets \vect{n}_\text{gen}$, \cref{eq:gen}\;    
    $f_\text{critic} \gets \vect{x}_\text{batch}, \vect{n}_\text{gen}, \vect{n}_\vect{h}$, \cref{eq:critic}\;  
}
\end{algorithm}

Figuratively, we mark all regions that are not filled by normal samples as anomalous, thus closing blind spots of the anomaly detector.
Summarising this chapter, we give a high-level overview about the entire training process in Algo.~\ref{alg:training}.
We also evaluated a simple anomaly generator, where we input $\vect{n}_\vect{h}$ directly to the decoder.

\begin{table*}[tb]
\caption{
Evaluated data sets along with the AAEs' hidden dimensions.
}\label{tab:data}
\centering
\begin{tabular}{c c c c c c c c}
\textbf{Name} & & \textbf{Normal}   & \textbf{Anomalous} & \textbf{Encoder} & \textbf{$\abs{\set{X}_\text{train, norm}}$} & \textbf{$\abs{\set{X}_\text{test, norm}}$} & \textbf{$\abs{\set{X}_\text{test, anom}}$} \\
\midrule
CC & \cite{pozzolo_calibrating_2015} & Normal & Anomalous & 50, 40, 30, 20, 10, 5 & \num{2.30e+05} & \num{4.26e+04} & \num{7.40e+01} \\
Census & \cite{dua_uci_2017} & $<50$k & $>50$k & 600, 300, 150, 75, 30, 15 & \num{8.89e+04} & \num{9.36e+04} & \num{6.19e+03}\\
CoverT & \cite{blackard_comparative_1999} & 1-4 & 5-7 & 90, 75, 60, 45, 25, 15 & \num{4.31e+05} & \num{8.01e+04} & \num{7.07e+03} \\
DoH & \cite{montazerishatoori_detection_2020} & Benign & Malicious & 50, 40, 30, 20, 10, 2 & \num{1.60e+04} & \num{2.94e+03} & \num{3.75e+04} \\
EMNIST & \cite{cohen_emnist_2017} & A-M & N-Z & 32C3-16C3-8C3-16C3 & \num{5.93e+04} & \num{1.04e+04} & \num{1.04e+04} \\
FMNIST & \cite{xiao_fashion-mnist_2017} & 0-4 & 5-9 & 32C3-16C3-8C3-16C3 & \num{2.85e+04} & \num{5.00e+03} & \num{5.00e+03} \\
KDD & \cite{tavallaee_detailed_2009} & Normal & Anomalous & 150, 100, 70, 40, 25, 10 & \num{6.39e+04} & \num{9.71e+03} & \num{1.28e+04} \\
MNIST & \cite{lecun_gradient-based_1998} & 0-4 & 5-9 & 32C3-16C3-8C3-16C3 & \num{2.91e+04} & \num{5.14e+03} & \num{4.86e+03} \\
Mammo. & \cite{woods_comparative_1993} & Normal & Malignant & 12, 10, 8, 6, 3, 2 & \num{8.82e+03} & \num{1.64e+03} & \num{3.90e+01} \\
URL & \cite{mamun_detecting_2016} & Benign & Def.-Spam & 100, 80, 60, 40, 20, 10 & \num{6.32e+03} & \num{1.14e+03} & \num{4.37e+03} \\
\end{tabular}
\end{table*}

\section{Experimental Setup}
\subsection{Parameter Choices}
\SAD{} comprises multiple NNs, e.g. the AAE and the discriminator, which require adequate layer sizes.
For simplicity, we chose the very same hyperparameters across all data sets except for the sizes of the AAE.
We did not perform exhaustive parameter tuning, but tried to match the layer sizes to the dimension of the input:
the middle layer of the AAE should be smaller than in input to compress the input features.
All AAEs are symmetric, i.e. the decoder dimensions are the same as the encoder dimensions but reversely arranged.
The discriminator had the same dimensions for all data sets, 40, 20, 10, 5, whereas the generator was 50, 40, 30, 20.
The alarm network uses 1000, 500, 200, 75, except for the low dimensional Mammography \cite{woods_comparative_1993} data set, where it was set to 100, 50, 25, 10.
We used Adam \cite{kingma_adam_2017} with a learning rate of $10^{-4}$ as optimiser for all networks.
As recommended in the Wasserstein GAN paper \cite{arjovsky_wasserstein_2017}, we apply weight clipping with a maximum norm of $0.01$.
The hidden layers are activated by Leaky ReLUs.
Dropout of 10\% is between each layer.
\SAD{} was trained for 500 epochs.

\subsection{Data Sets}
We chose ten publicly available data sets as mixture of common ML baseline data sets as well as important AD applications, e.g. fraud and intrusion detection.
We scaled the input data to the range $[0,1]$.
Categorical labels were 1-Hot encoded.
Samples, which did not contain numerical values afterwards, were dropped.
We used 80\% of the data in the training, 5\% in the validation and 15\% in the test set in case no test set was provided.

\subsection{Baseline Methods}
\label{sec:baselines}
\SAD{} is an unsupervised DL-\&GAN-based AD method.
We chose six state-of-the-art baseline AD methods of related categories.
Deep-SVDD \cite{ruff_deep_2018} is a DL-based one-class classifier, DAGMM \cite{zong_deep_2018} combines AEs and Gaussian mixture models.
f-AnoGAN \cite{schlegl_f-anogan_2019} is a commonly used GAN-based AD baseline.
GANomaly \cite{akcay_ganomaly_2019} is a GAN-based AD method featuring an AAE-like architecture, FenceGAN \cite{ngo_fence_2019} encloses the training data by generated anomalies.
REPEN \cite{pang_learning_2018} is an outlier detection (OD) method with DL feature scaling.
Moreover, we evaluated the impact of our novel anomaly generator in an ablation study with \AAA{} \cite{sperl_activation_2021}.
Whenever applicable, we used the very same architecture for the baseline methods as for \SAD{} to allow a fair comparison.
If the authors evaluated the respective data set, we favoured their architecture.

\section{Evaluation}
We show our detection performance compared to the baseline methods with the area under the ROC curve (AUC) as metric.
The AUC is commonly used in AD (e.g. \cite{ruff_deep_2018,pang_deep_2019,sperl_activation_2021}), measuring the trade-off between the true and false positive rate independent of a detection threshold.
An ideal AD method has an AUC of~1.
For a complementary view, we also report the average precision (AP).
We refer to the discussion of Hendrycks \& Gimpel \cite{hendrycks_baseline_2017} about the advantages of these two metrics in the setting of AD.

\begin{figure*}[tb]
  \begin{center}
    \input{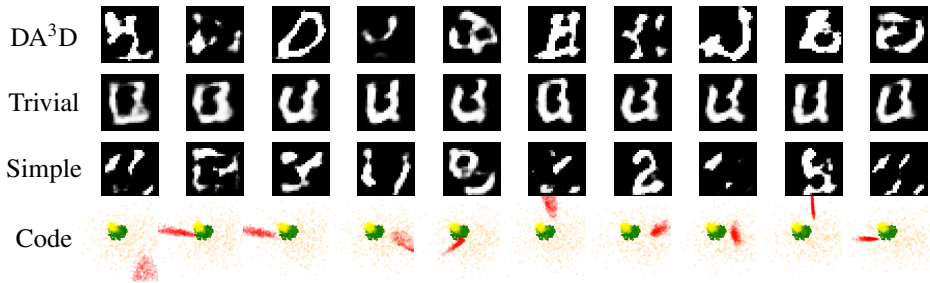}
    \caption{
        Synthetic anomalies from epoch 50 to 500 for MNIST, where the digit 0-4 were known during training.
        In the code space, we show the normal data (green) as well as trivial (yellow), simple (orange) and generated (red) anomalies.
    }
    \label{fig:generated}
  \end{center}
\end{figure*}

\subsection{Visual Inspection}
We start our evaluation by a visual inspection of the generated anomalies shown in \Cref{fig:generated}.
It is instructive to see if they match our expectations, although we can judge the quality of the generated samples only by their impact on the anomaly detection performance.
As comparison, we show the trivial and simple anomalies, i.e. input and latent space anomalies sampled from a Gaussian prior.
Looking at the anomalies generated by \SAD{}, we see two important characteristics:
They 1)~are different to the normal distribution, 2)~exhibit some variance, i.e. do not imitate previously generated data.
In contrast, some of the simple anomalies are visually close to the training data, e.g. the digit~3 and~2.
For the trivial anomalies, the output is a mixture of shapes seen in the normal data, yet stationary in form.
In our ablation study, we will explore the performance when only using these anomalies.
Analysing the latent space, we see \SAD{} to sample around the green space of normal samples.
Generally, our two adversarial goals seem to be useful to sample outside the vicinity of the normal yet close to the normal space.
In some epochs, the generated samples overlapped the normal samples, yet not in the extend of the trivial and simple anomalies.
In the following chapters, we analyse if the generated anomalies are indeed useful to boost the AD performance.

\begin{table*}[tb]
\caption{Test results as mean AUC and AP along with the standard deviation after six runs.}
\label{exp:results}
\centering
\setlength{\tabcolsep}{1pt}
\resizebox{\textwidth}{!}{%

\begin{tabular}{>{\color{gray}}c >{\color{gray}}c >{\color{gray}}c >{\color{gray}}c >{\color{gray}}c >{\color{gray}}c | >{\color{gray}}c >{\color{gray}}c >{\color{gray}}c >{\color{gray}}c >{\color{gray}}c >{\color{gray}}c >{\color{gray}}c >{\color{gray}}c >{\color{gray}}c >{\color{gray}}c >{\color{gray}}c >{\color{gray}}c | >{\color{gray}}c >{\color{gray}}c }
\toprule
& & \multicolumn{4}{>{\color{gray}}c }{Ours} & \multicolumn{12}{>{\color{gray}}c }{Baselines} & \multicolumn{2}{>{\color{gray}}c }{Ablation}\\

     &  & \multicolumn{2}{c}{DA3D} & \multicolumn{2}{c}{Simple} & \multicolumn{2}{c}{DeepSVDD} & \multicolumn{2}{c}{DAGMM} & \multicolumn{2}{c}{REPEN} & \multicolumn{2}{c}{fAnoGAN} & \multicolumn{2}{c}{GANomaly} & \multicolumn{2}{c}{FenceGAN} & \multicolumn{2}{c}{A3} \\
Poll.     &  &                                             AUC &                                              AP &                                             AUC &                                              AP &                                AUC &                                 AP &                                             AUC &                                              AP &                                             AUC &                                              AP &                                AUC &                                 AP &                                             AUC &                                              AP &                                AUC &                                              AP &                                             AUC &                                              AP \\
\midrule
0\% & CC &               $.90\scriptscriptstyle \pm .03$ &  \color{black}$.43\scriptscriptstyle \pm .15$ &               $.88\scriptscriptstyle \pm .06$ &               $.38\scriptscriptstyle \pm .18$ &  $.89\scriptscriptstyle \pm .02$ &  $.38\scriptscriptstyle \pm .11$ &               $.84\scriptscriptstyle \pm .02$ &               $.02\scriptscriptstyle \pm .01$ &  \color{black}$.92\scriptscriptstyle \pm .02$ &               $.41\scriptscriptstyle \pm .11$ &  $.87\scriptscriptstyle \pm .04$ &  $.18\scriptscriptstyle \pm .05$ &               $.84\scriptscriptstyle \pm .04$ &               $.29\scriptscriptstyle \pm .09$ &  $.53\scriptscriptstyle \pm .29$ &               $.13\scriptscriptstyle \pm .16$ &               $.87\scriptscriptstyle \pm .02$ &               $.25\scriptscriptstyle \pm .05$ \\
     & Census &  \color{black}$.67\scriptscriptstyle \pm .05$ &               $.11\scriptscriptstyle \pm .02$ &  \color{black}$.67\scriptscriptstyle \pm .03$ &  \color{black}$.13\scriptscriptstyle \pm .03$ &  $.59\scriptscriptstyle \pm .02$ &  $.09\scriptscriptstyle \pm .01$ &               $.40\scriptscriptstyle \pm .08$ &               $.05\scriptscriptstyle \pm .01$ &               $.63\scriptscriptstyle \pm .02$ &               $.08\scriptscriptstyle \pm .00$ &  $.53\scriptscriptstyle \pm .11$ &  $.08\scriptscriptstyle \pm .03$ &               $.53\scriptscriptstyle \pm .07$ &               $.07\scriptscriptstyle \pm .01$ &  $.60\scriptscriptstyle \pm .09$ &               $.09\scriptscriptstyle \pm .02$ &               $.65\scriptscriptstyle \pm .06$ &               $.09\scriptscriptstyle \pm .01$ \\
     & CoverT &               $.69\scriptscriptstyle \pm .05$ &               $.16\scriptscriptstyle \pm .03$ &               $.63\scriptscriptstyle \pm .07$ &               $.15\scriptscriptstyle \pm .02$ &  $.54\scriptscriptstyle \pm .02$ &  $.11\scriptscriptstyle \pm .01$ &  \color{black}$.73\scriptscriptstyle \pm .04$ &  \color{black}$.21\scriptscriptstyle \pm .04$ &               $.71\scriptscriptstyle \pm .04$ &               $.16\scriptscriptstyle \pm .02$ &  $.61\scriptscriptstyle \pm .03$ &  $.14\scriptscriptstyle \pm .03$ &               $.69\scriptscriptstyle \pm .05$ &               $.17\scriptscriptstyle \pm .04$ &  $.47\scriptscriptstyle \pm .14$ &               $.08\scriptscriptstyle \pm .03$ &               $.49\scriptscriptstyle \pm .09$ &               $.10\scriptscriptstyle \pm .02$ \\
     & DoH &  \color{black}$.95\scriptscriptstyle \pm .05$ &  \color{black}$.99\scriptscriptstyle \pm .01$ &               $.93\scriptscriptstyle \pm .08$ &  \color{black}$.99\scriptscriptstyle \pm .01$ &  $.76\scriptscriptstyle \pm .11$ &  $.98\scriptscriptstyle \pm .01$ &               $.67\scriptscriptstyle \pm .08$ &               $.96\scriptscriptstyle \pm .01$ &               $.25\scriptscriptstyle \pm .03$ &               $.89\scriptscriptstyle \pm .01$ &  $.64\scriptscriptstyle \pm .16$ &  $.95\scriptscriptstyle \pm .02$ &               $.61\scriptscriptstyle \pm .14$ &               $.95\scriptscriptstyle \pm .02$ &  $.47\scriptscriptstyle \pm .14$ &               $.91\scriptscriptstyle \pm .03$ &               $.70\scriptscriptstyle \pm .13$ &               $.96\scriptscriptstyle \pm .02$ \\
     & EMNIST &  \color{black}$.69\scriptscriptstyle \pm .06$ &  \color{black}$.69\scriptscriptstyle \pm .06$ &               $.68\scriptscriptstyle \pm .06$ &               $.68\scriptscriptstyle \pm .06$ &  $.56\scriptscriptstyle \pm .01$ &  $.57\scriptscriptstyle \pm .01$ &               $.53\scriptscriptstyle \pm .01$ &               $.55\scriptscriptstyle \pm .01$ &               $.54\scriptscriptstyle \pm .03$ &               $.53\scriptscriptstyle \pm .03$ &  $.59\scriptscriptstyle \pm .04$ &  $.56\scriptscriptstyle \pm .04$ &               $.62\scriptscriptstyle \pm .04$ &               $.58\scriptscriptstyle \pm .02$ &  $.54\scriptscriptstyle \pm .00$ &               $.53\scriptscriptstyle \pm .00$ &               $.60\scriptscriptstyle \pm .02$ &               $.58\scriptscriptstyle \pm .01$ \\
     & FMNIST &               $.90\scriptscriptstyle \pm .02$ &               $.92\scriptscriptstyle \pm .02$ &  \color{black}$.92\scriptscriptstyle \pm .01$ &  \color{black}$.94\scriptscriptstyle \pm .01$ &  $.74\scriptscriptstyle \pm .02$ &  $.77\scriptscriptstyle \pm .03$ &               $.84\scriptscriptstyle \pm .03$ &               $.80\scriptscriptstyle \pm .04$ &               $.62\scriptscriptstyle \pm .04$ &               $.66\scriptscriptstyle \pm .05$ &  $.80\scriptscriptstyle \pm .08$ &  $.79\scriptscriptstyle \pm .11$ &               $.87\scriptscriptstyle \pm .02$ &               $.90\scriptscriptstyle \pm .01$ &  $.84\scriptscriptstyle \pm .00$ &               $.88\scriptscriptstyle \pm .00$ &               $.87\scriptscriptstyle \pm .04$ &               $.85\scriptscriptstyle \pm .05$ \\
     & KDD &               $.86\scriptscriptstyle \pm .03$ &               $.85\scriptscriptstyle \pm .04$ &  \color{black}$.90\scriptscriptstyle \pm .01$ &               $.87\scriptscriptstyle \pm .04$ &  $.81\scriptscriptstyle \pm .07$ &  $.86\scriptscriptstyle \pm .05$ &               $.88\scriptscriptstyle \pm .03$ &               $.89\scriptscriptstyle \pm .03$ &               $.83\scriptscriptstyle \pm .02$ &               $.83\scriptscriptstyle \pm .02$ &  $.63\scriptscriptstyle \pm .17$ &  $.72\scriptscriptstyle \pm .11$ &               $.89\scriptscriptstyle \pm .03$ &  \color{black}$.90\scriptscriptstyle \pm .03$ &  $.72\scriptscriptstyle \pm .25$ &               $.81\scriptscriptstyle \pm .17$ &               $.73\scriptscriptstyle \pm .13$ &               $.76\scriptscriptstyle \pm .12$ \\
     & MNIST &               $.67\scriptscriptstyle \pm .10$ &               $.63\scriptscriptstyle \pm .10$ &               $.56\scriptscriptstyle \pm .08$ &               $.55\scriptscriptstyle \pm .07$ &  $.59\scriptscriptstyle \pm .01$ &  $.57\scriptscriptstyle \pm .01$ &               $.74\scriptscriptstyle \pm .01$ &  \color{black}$.74\scriptscriptstyle \pm .01$ &               $.49\scriptscriptstyle \pm .02$ &               $.44\scriptscriptstyle \pm .01$ &  $.56\scriptscriptstyle \pm .09$ &  $.52\scriptscriptstyle \pm .09$ &  \color{black}$.79\scriptscriptstyle \pm .07$ &  \color{black}$.74\scriptscriptstyle \pm .08$ &  $.62\scriptscriptstyle \pm .02$ &               $.59\scriptscriptstyle \pm .01$ &               $.42\scriptscriptstyle \pm .01$ &               $.42\scriptscriptstyle \pm .01$ \\
     & Mammo. &               $.84\scriptscriptstyle \pm .03$ &               $.38\scriptscriptstyle \pm .09$ &  \color{black}$.86\scriptscriptstyle \pm .03$ &  \color{black}$.40\scriptscriptstyle \pm .11$ &  $.68\scriptscriptstyle \pm .05$ &  $.07\scriptscriptstyle \pm .02$ &               $.80\scriptscriptstyle \pm .05$ &               $.15\scriptscriptstyle \pm .12$ &  \color{black}$.86\scriptscriptstyle \pm .01$ &               $.29\scriptscriptstyle \pm .03$ &  $.66\scriptscriptstyle \pm .14$ &  $.15\scriptscriptstyle \pm .08$ &               $.71\scriptscriptstyle \pm .29$ &               $.31\scriptscriptstyle \pm .28$ &  $.66\scriptscriptstyle \pm .22$ &               $.19\scriptscriptstyle \pm .22$ &               $.80\scriptscriptstyle \pm .12$ &               $.33\scriptscriptstyle \pm .05$ \\
     & URL &  \color{black}$.89\scriptscriptstyle \pm .02$ &  \color{black}$.97\scriptscriptstyle \pm .01$ &               $.88\scriptscriptstyle \pm .01$ &  \color{black}$.97\scriptscriptstyle \pm .00$ &  $.83\scriptscriptstyle \pm .02$ &  $.95\scriptscriptstyle \pm .00$ &               $.74\scriptscriptstyle \pm .02$ &               $.91\scriptscriptstyle \pm .01$ &               $.80\scriptscriptstyle \pm .03$ &               $.93\scriptscriptstyle \pm .01$ &  $.83\scriptscriptstyle \pm .01$ &  $.94\scriptscriptstyle \pm .01$ &               $.86\scriptscriptstyle \pm .02$ &               $.96\scriptscriptstyle \pm .01$ &  $.66\scriptscriptstyle \pm .18$ &               $.89\scriptscriptstyle \pm .08$ &               $.88\scriptscriptstyle \pm .02$ &  \color{black}$.97\scriptscriptstyle \pm .01$ \\
     & mean &                             \color{black}$.81$ &                             \color{black}$.61$ &                                          $.79$ &                             \color{black}$.61$ &                             $.70$ &                             $.54$ &                                          $.72$ &                                          $.53$ &                                          $.67$ &                                          $.52$ &                             $.67$ &                             $.50$ &                                          $.74$ &                                          $.59$ &                             $.61$ &                                          $.51$ &                                          $.70$ &                                          $.53$ \\

\midrule

1\% & CC &               $.86\scriptscriptstyle \pm .07$ &               $.19\scriptscriptstyle \pm .07$ &               $.87\scriptscriptstyle \pm .05$ &               $.22\scriptscriptstyle \pm .08$ &  $.87\scriptscriptstyle \pm .02$ &  $.12\scriptscriptstyle \pm .06$ &               $.86\scriptscriptstyle \pm .04$ &               $.02\scriptscriptstyle \pm .03$ &  \color{black}$.91\scriptscriptstyle \pm .01$ &  \color{black}$.27\scriptscriptstyle \pm .06$ &  $.76\scriptscriptstyle \pm .31$ &  $.17\scriptscriptstyle \pm .16$ &               $.70\scriptscriptstyle \pm .06$ &               $.07\scriptscriptstyle \pm .07$ &  $.66\scriptscriptstyle \pm .20$ &               $.06\scriptscriptstyle \pm .04$ &               $.89\scriptscriptstyle \pm .02$ &               $.16\scriptscriptstyle \pm .09$ \\
     & Census &               $.56\scriptscriptstyle \pm .04$ &               $.07\scriptscriptstyle \pm .01$ &               $.61\scriptscriptstyle \pm .06$ &  \color{black}$.10\scriptscriptstyle \pm .03$ &  $.55\scriptscriptstyle \pm .02$ &  $.08\scriptscriptstyle \pm .00$ &               $.38\scriptscriptstyle \pm .08$ &               $.05\scriptscriptstyle \pm .01$ &  \color{black}$.66\scriptscriptstyle \pm .02$ &               $.08\scriptscriptstyle \pm .00$ &  $.53\scriptscriptstyle \pm .04$ &  $.07\scriptscriptstyle \pm .01$ &               $.57\scriptscriptstyle \pm .07$ &               $.08\scriptscriptstyle \pm .01$ &  $.48\scriptscriptstyle \pm .06$ &               $.06\scriptscriptstyle \pm .01$ &               $.64\scriptscriptstyle \pm .06$ &               $.08\scriptscriptstyle \pm .01$ \\
     & CoverT &               $.64\scriptscriptstyle \pm .07$ &               $.15\scriptscriptstyle \pm .03$ &               $.60\scriptscriptstyle \pm .05$ &               $.12\scriptscriptstyle \pm .01$ &  $.58\scriptscriptstyle \pm .03$ &  $.12\scriptscriptstyle \pm .02$ &  \color{black}$.73\scriptscriptstyle \pm .03$ &  \color{black}$.20\scriptscriptstyle \pm .03$ &               $.72\scriptscriptstyle \pm .02$ &               $.17\scriptscriptstyle \pm .02$ &  $.54\scriptscriptstyle \pm .07$ &  $.11\scriptscriptstyle \pm .01$ &               $.66\scriptscriptstyle \pm .06$ &               $.16\scriptscriptstyle \pm .03$ &  $.56\scriptscriptstyle \pm .10$ &               $.11\scriptscriptstyle \pm .03$ &               $.34\scriptscriptstyle \pm .12$ &               $.07\scriptscriptstyle \pm .02$ \\
     & DoH &  \color{black}$.81\scriptscriptstyle \pm .05$ &  \color{black}$.98\scriptscriptstyle \pm .00$ &               $.78\scriptscriptstyle \pm .11$ &  \color{black}$.98\scriptscriptstyle \pm .01$ &  $.60\scriptscriptstyle \pm .06$ &  $.95\scriptscriptstyle \pm .01$ &               $.66\scriptscriptstyle \pm .06$ &               $.96\scriptscriptstyle \pm .01$ &               $.15\scriptscriptstyle \pm .05$ &               $.86\scriptscriptstyle \pm .02$ &  $.39\scriptscriptstyle \pm .06$ &  $.90\scriptscriptstyle \pm .02$ &               $.65\scriptscriptstyle \pm .09$ &               $.95\scriptscriptstyle \pm .01$ &  $.55\scriptscriptstyle \pm .18$ &               $.93\scriptscriptstyle \pm .04$ &               $.80\scriptscriptstyle \pm .18$ &               $.97\scriptscriptstyle \pm .03$ \\
     & EMNIST &  \color{black}$.63\scriptscriptstyle \pm .03$ &  \color{black}$.63\scriptscriptstyle \pm .04$ &               $.62\scriptscriptstyle \pm .05$ &               $.62\scriptscriptstyle \pm .05$ &  $.54\scriptscriptstyle \pm .02$ &  $.55\scriptscriptstyle \pm .02$ &               $.53\scriptscriptstyle \pm .01$ &               $.54\scriptscriptstyle \pm .01$ &               $.54\scriptscriptstyle \pm .02$ &               $.53\scriptscriptstyle \pm .01$ &  $.59\scriptscriptstyle \pm .05$ &  $.56\scriptscriptstyle \pm .04$ &               $.61\scriptscriptstyle \pm .02$ &               $.58\scriptscriptstyle \pm .01$ &  $.54\scriptscriptstyle \pm .01$ &               $.53\scriptscriptstyle \pm .01$ &               $.60\scriptscriptstyle \pm .01$ &               $.58\scriptscriptstyle \pm .01$ \\
     & FMNIST &               $.74\scriptscriptstyle \pm .10$ &               $.73\scriptscriptstyle \pm .09$ &               $.75\scriptscriptstyle \pm .08$ &               $.73\scriptscriptstyle \pm .06$ &  $.68\scriptscriptstyle \pm .01$ &  $.69\scriptscriptstyle \pm .01$ &               $.82\scriptscriptstyle \pm .03$ &               $.76\scriptscriptstyle \pm .04$ &               $.75\scriptscriptstyle \pm .07$ &               $.68\scriptscriptstyle \pm .06$ &  $.80\scriptscriptstyle \pm .10$ &  $.81\scriptscriptstyle \pm .10$ &               $.82\scriptscriptstyle \pm .02$ &               $.82\scriptscriptstyle \pm .03$ &  $.79\scriptscriptstyle \pm .01$ &  \color{black}$.83\scriptscriptstyle \pm .01$ &  \color{black}$.86\scriptscriptstyle \pm .02$ &               $.80\scriptscriptstyle \pm .03$ \\
     & KDD &               $.83\scriptscriptstyle \pm .09$ &               $.81\scriptscriptstyle \pm .10$ &               $.87\scriptscriptstyle \pm .04$ &               $.85\scriptscriptstyle \pm .05$ &  $.79\scriptscriptstyle \pm .08$ &  $.84\scriptscriptstyle \pm .05$ &  \color{black}$.89\scriptscriptstyle \pm .03$ &               $.89\scriptscriptstyle \pm .03$ &               $.87\scriptscriptstyle \pm .03$ &               $.88\scriptscriptstyle \pm .03$ &  $.68\scriptscriptstyle \pm .21$ &  $.76\scriptscriptstyle \pm .16$ &               $.88\scriptscriptstyle \pm .07$ &  \color{black}$.90\scriptscriptstyle \pm .05$ &  $.74\scriptscriptstyle \pm .10$ &               $.82\scriptscriptstyle \pm .05$ &               $.67\scriptscriptstyle \pm .14$ &               $.68\scriptscriptstyle \pm .11$ \\
     & MNIST &               $.74\scriptscriptstyle \pm .06$ &               $.72\scriptscriptstyle \pm .06$ &               $.60\scriptscriptstyle \pm .13$ &               $.61\scriptscriptstyle \pm .11$ &  $.57\scriptscriptstyle \pm .01$ &  $.56\scriptscriptstyle \pm .01$ &               $.71\scriptscriptstyle \pm .02$ &               $.72\scriptscriptstyle \pm .02$ &               $.48\scriptscriptstyle \pm .01$ &               $.45\scriptscriptstyle \pm .01$ &  $.53\scriptscriptstyle \pm .06$ &  $.50\scriptscriptstyle \pm .04$ &  \color{black}$.80\scriptscriptstyle \pm .03$ &  \color{black}$.76\scriptscriptstyle \pm .04$ &  $.61\scriptscriptstyle \pm .01$ &               $.58\scriptscriptstyle \pm .01$ &               $.44\scriptscriptstyle \pm .03$ &               $.43\scriptscriptstyle \pm .03$ \\
     & Mammo. &               $.79\scriptscriptstyle \pm .08$ &               $.16\scriptscriptstyle \pm .10$ &               $.80\scriptscriptstyle \pm .07$ &  \color{black}$.21\scriptscriptstyle \pm .13$ &  $.58\scriptscriptstyle \pm .02$ &  $.03\scriptscriptstyle \pm .00$ &               $.78\scriptscriptstyle \pm .04$ &               $.13\scriptscriptstyle \pm .07$ &  \color{black}$.82\scriptscriptstyle \pm .03$ &               $.13\scriptscriptstyle \pm .06$ &  $.62\scriptscriptstyle \pm .22$ &  $.14\scriptscriptstyle \pm .08$ &               $.68\scriptscriptstyle \pm .23$ &               $.13\scriptscriptstyle \pm .16$ &  $.60\scriptscriptstyle \pm .22$ &               $.07\scriptscriptstyle \pm .08$ &               $.61\scriptscriptstyle \pm .17$ &               $.14\scriptscriptstyle \pm .07$ \\
     & URL &               $.85\scriptscriptstyle \pm .06$ &               $.95\scriptscriptstyle \pm .03$ &  \color{black}$.86\scriptscriptstyle \pm .03$ &  \color{black}$.96\scriptscriptstyle \pm .01$ &  $.75\scriptscriptstyle \pm .02$ &  $.93\scriptscriptstyle \pm .01$ &               $.75\scriptscriptstyle \pm .02$ &               $.91\scriptscriptstyle \pm .01$ &               $.72\scriptscriptstyle \pm .02$ &               $.89\scriptscriptstyle \pm .01$ &  $.81\scriptscriptstyle \pm .02$ &  $.93\scriptscriptstyle \pm .01$ &               $.80\scriptscriptstyle \pm .06$ &               $.94\scriptscriptstyle \pm .03$ &  $.69\scriptscriptstyle \pm .11$ &               $.90\scriptscriptstyle \pm .04$ &               $.82\scriptscriptstyle \pm .05$ &               $.94\scriptscriptstyle \pm .01$ \\
     & mean &                             \color{black}$.74$ &                             \color{black}$.54$ &                             \color{black}$.74$ &                             \color{black}$.54$ &                             $.65$ &                             $.49$ &                                          $.71$ &                                          $.52$ &                                          $.66$ &                                          $.49$ &                             $.62$ &                             $.50$ &                                          $.72$ &                             \color{black}$.54$ &                             $.62$ &                                          $.49$ &                                          $.67$ &                                          $.48$ \\
\bottomrule
\end{tabular}

}
\end{table*}

\subsection{Ideal Performance}
In the upper part of \Cref{exp:results}, we show the detection results under ideal conditions, i.e. a training data set, where all samples are known to be normal.
We are happy to report that \SAD{} scored the best with a mean AUC of $81\%$.
This is a \percincrease{.74}{.81}increase to the next best baseline method, GANomaly.
Also the simple anomaly generator based on Gaussian noise in the latent space performed remarkably well, on average only \percincrease{.81}{.79}worse than the GAN-based one.
In total, the advanced anomaly generator was better or equal on 6 of 10 experiments, most significantly on CoverT and MNIST.
The high-dimensional code space distribution may be sparse enough that the random samples do not significantly increase the false positive rate.
We conclude that our main assumption holds true, i.e. we can automatically generate synthetic anomalies from the latent space of AAEs, which are useful to detect real anomalies.

\SAD{} took the lead across many different types of data, e.g. on tabular data as DoH (\percincrease{.76}{.95}better than DeepSVDD) or image data as EMNIST (\percincrease{.62}{.69}better than GANomaly).
Even on anonymised data as in CC, where there is no known relation between the input features, \SAD{} was only \percincrease{.92}{.9}worse than REPEN.
To underline the general performance of \SAD{}, we kept the very same set of hyperparameters for all data sets under evaluation.
In our ablation study, \Cref{chap:ablation}, we evaluated the influence of our main hyperparameter, i.e. the sampling standard deviation.
In total, both versions of our anomaly generator beat the baseline methods on 7~out of~10 data sets.
With \SAD{}, we provide a flexible unsupervised AD method applicable to a wide range of use cases.

\subsection{Noise Resistance}
In the next set of experiments, summarised by the lower half of \Cref{exp:results}, we assessed \SAD{}'s robustness against polluted training data.
Here, 1\% of the training data were of anomalous nature while being processed as normal.
This scenario is often found in practice, where some unknown anomalies are within the training data, yet the majority is normal.
Generally, all GAN-based AD methods exhibited a higher variance, which may be explained by the stochastic training process paired with the complex loss surface of the discriminator-generator networks.
Again, \SAD{} scored best with a mean AUC of $74\%$, on par with the simple anomaly generator and closely followed by GANomaly.
The performance remained stable on most data sets with a significant drop on FMNIST, Census and DoH.
Analysing FMNIST, the anomalous classes there (e.g. ``shirt'') feature shapes similar to the normal classes (e.g. ``t-shirt''), thus may be mapped close to the normal distribution.
However, \SAD{} ignores normal samples in the generation process, thus likely samples around these unknown anomalies.
A joined training of the AAE and \SAD{} may solve this issue in future.
Summarising this chapter, we saw \SAD{} to perform well on noisy data sets as found in real-world scenarios.

\subsection{Ablation Study}
\label{chap:ablation}
In the final set of experiments, we critically rethink the architecture of \SAD{}.

\begin{figure}[tb]
  \begin{center}
    \pgfplotstableread[col sep=comma]{csv/ablation.csv}\data
\pgfplotsset{error_p/.style={error bars/.cd,	y dir=both,y explicit relative}}

\begin{tikzpicture}
	\begin{axis}[
        height=4cm,
		width=\linewidth,
		symbolic x coords={Mammo., URL, KDD, CC, DoH, Census, CoverT, MNIST, FMNIST, EMNIST},
		xticklabel style={rotate=25},
		legend style={at={(0.45, -0.75)},anchor=north, legend columns=4, nodes={scale=0.7, transform shape}},
		grid=both,
		xtick=data,
		xlabel={Data Set},
		ylabel={Test AUC},
	]
		\addplot+[error_p] table[x=Exp.,y=DA3D-1-AUC-mean, y error=DA3D-1-AUC-std] {\data};
		\addplot+[error_p] table[x=Exp.,y=DA3D-AUC-mean, y error=DA3D-AUC-std] {\data};
		\addplot+[error_p] table[x=Exp.,y=DA3D-3-AUC-mean, y error=DA3D-3-AUC-std] {\data};
		\addplot+[error_p] table[x=Exp.,y=DA3D-4-AUC-mean, y error=DA3D-4-AUC-std] {\data};
		\legend{$\sigma=1$, $\sigma=2$, $\sigma=3$, $\sigma=4$}
	\end{axis}
\end{tikzpicture}
    \caption{
        Impact of the sampling standard deviation $\vect{n}_\vect{h}$ on the detection performance shown as mean AUC after 6 runs.
    }
    \label{fig:ablation}
  \end{center}
\end{figure}
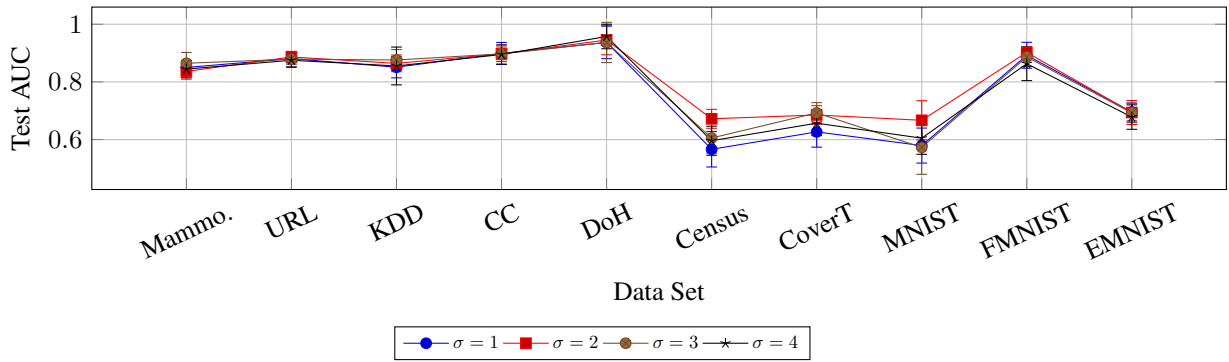

\paragraph{Influence of the Sampling Standard Deviation}
\label{exp:stddev}
So far, we reported \SAD{}'s performance using the very same sampling standard deviation $\sigma=2$.
Remember that the parameter influences the distance in the AAE's latent space, where the synthetic anomalies are sampled.
In our first ablation study, we analysed the impact of this parameter on the detection performance.
In \Cref{fig:ablation}, we show the mean performance for 4~different configurations.
We see that \SAD{}'s performance peaks at $\sigma=2$, with only minor performance degradation for the other settings.
The mean performance across all data sets for $\sigma=2$ was $81$\%, whereas the worst configurations ($\sigma=\{1,4\}$) were at $78\%$.
We conclude that \SAD{} performs well with the very same hyperparameters across all 10 data sets under evaluation.

\paragraph{Influence of the Advanced Synthetic Anomalies}
\AAA{} is the semi-supervised AD method, which we combined our anomaly generator with to form our unsupervised AD method \SAD{}.
\AAA{} only used Gaussian noise as synthetic anomalies, whereas we introduced advanced ones in \SAD{}.
We show the results of our second ablation study on the last column in \Cref{exp:results}.
Under clean training data, \SAD{} performed \percincrease{.7}{.81}better, beating \AAA{} on all data sets.
On polluted data, we performed \percincrease{.67}{.74}better.
\AAA{}'s trivial anomalies alone seemed not enough for adequate unsupervised performance.
We conclude that the advanced anomalies generated in \SAD{} significantly helped \AAA{} to generalise to real anomalies.

\section*{Discussion and Future Work}
\SAD{} introduces as novel GAN-based anomaly generator, leveraging a semi-supervised AD method to the considerably stricter unsupervised setting.
We hope to spark interest to integrate a generator feedback-loop to other semi-supervised AD methods.
GANs profit from an active research community.
In our research, we incorporated recent advances like optimised loss functions.
We are certain that future research on GANs will also improve the performance of our anomaly generator.
Thus, we like to encourage to update and rethink the building blocks used in \SAD{} as the research on generative models progresses.

\subsection*{Conclusion}
In this paper, we introduced \SAD{}: an unsupervised anomaly detection method combining generative adversarial nets and the analysis of hidden activation values.
By generating anomalous counterexamples from normal samples only, we leveraged a semi-supervised AD method to the strict and more widely applicable setting of unsupervised AD.
In other words, we showed that an AD method trained on our artificial anomalies sampled from the latent space of an AAE is able to detect real anomalies.
The anomaly detector and the anomaly generator are interconnected, challenging themselves the clearer the notion of normal samples becomes.
\SAD{} works in a purely data-driven way, thus can be applied on a variety of use cases without any domain expert required.

\bibliographystyle{abbrv}
\bibliography{references}

\end{document}